\pdfoutput=1

\documentclass[11pt]{article}

\usepackage[preprint]{acl}

\usepackage{times}
\usepackage{latexsym}

\usepackage[T1]{fontenc}

\usepackage[utf8]{inputenc}

\usepackage{microtype}

\usepackage{inconsolata}

\usepackage{graphicx}
\usepackage{booktabs}

\usepackage{verbatim}  
\usepackage{verbatimbox}
\usepackage{spverbatim}

\usepackage{multirow}

\usepackage{listings}
\title{OpenFActScore: Open-Source Atomic Evaluation of Factuality in Text Generation}

  \author{Lucas Fonseca Lage$^{1,4}$ \and Simon Ostermann$^{2,3,4}$   \\
$^1$Philipps-Universität Marburg \\
$^2$German Research Centre for Artificial Intelligence (DFKI) \\ 
$^3$Centre for European Research in Trusted AI (CERTAIN) \\
$^4$Saarland University \\
\texttt{lage@uni-marburg.de}\qquad\texttt{simon.ostermann@dfki.de}
}

\begin{document}
\maketitle
\begin{abstract}

We introduce OpenFActScore, an open-source implementation of the FActScore framework for evaluating the factuality of text generated by large language models (LLMs). FActScore evaluates the factual accuracy of long-form text by using Atomic Fact Generation (AFG) to extract individual factual claims and Atomic Fact Validation (AFV) to verify each claim against a trusted knowledge source.
While the original FActScore relies on closed-source and commercial models such as InstructGPT and ChatGPT, OpenFActScore enables the use of any Hugging Face-compatible model for both AFG and AFV.
We provide a detailed technical overview of our implementation, highlighting design choices and modifications made to support open models. 
We evaluate multiple open-source LLMs on both AFG and AFV using the original FActScore benchmark, reporting BERTScore-F1 for AFG and Error Rate relative to human annotations for AFV.
Our results show that open models can approximate the performance of closed-source systems, with Gemma achieving the best overall performance and our final setupt obtains a 0.99 Pearson correlation with the original FActScore experiments.
OpenFActScore promotes transparency, reproducibility, and cost-effective evaluation, and is available at: \url{https://github.com/lflage/OpenFActScore}.
\end{abstract}

\section{Introduction}

With the increasing popularity of Large Language Models (LLMs) for all kinds of everyday tasks, the need for a holistic evaluation of the output of such models increases.
The evaluation of LLMs is often performed across a wide range of tasks. Prominent examples include reasoning \cite{mondorfbeyond}, mathematics and arithmetics \cite{ahn-etal-2024-large}, ethical alignment \cite{huang2023trustgpt}, and more classical tasks such as question answering \cite{rajpurkar-etal-2016-squad} or natural language inference \cite{N18-1101} (a comprehensive overview can be found in \citet{guo2023evaluating}). 

For many cases, a simple gold-standard-based evaluation of model performance is conducted, using measures such as accuracy or F-Score. However, for evaluating some dimensions of model output, creating a gold standard is not trivial. One such dimension concerns \textit{factuality}, i.e. whether each facet of the generated output is factually correct. To evaluate factuality of an open output text, model-based evaluation is sometimes employed. As one of the most prominent examples, \citet{min2023-factscore} introduced \textit{FActScore}, a model-based metric of factuality. 

FActScore essentially formulates the evaluation of factuality as a 2-stage process, consisting of Atomic Fact Generation (AFG), to extract individual factual claims, and Atomic Fact Validation (AFV), to verify each claim against a trusted knowledge source. Factuality is then defined as the number of atomic facts, the smallest information units in a sentence, in the model output that are supported by the knowledge source. As a means to evaluate FActScore's performance, the method is initially manually evaluated with human annotators. To reduce costs and make the model scale better, the authors then create a model-based FActScore estimator using both InstructGPT \cite{ouyang2022training} and ChatGPT to automate the human annotator' work.
These estimators, although cheaper than using human annotators, are dependent on closed source models, which are also expensive depending on how many models are evaluated.

To alleviate this shortcoming, we introduce \textit{OpenFActScore}, an effort for fully open sourcing the original FActScore method. OpenFActScore is a straightforward extension of FActScore, providing access to any model available on Huggingface as estimators for FActScore.


The contributions of this paper are as follows:
\begin{itemize}
    \item We introduce OpenFActScore, an open-source variant of the widely-used model evaluation framework FActscore, for evaluating the factual correctness of model outputs.
    \item We provide a thorough technical description of the framework, highlighting differences and similarities to the original FActScore for full comparability.
    \item We evaluate OpenFActscore using the human annotations provided by \citet{min2023-factscore} to compare its performance to the original implementation.
\end{itemize}


In the following sections, we provide an in-depth description of the original FActScore (Section \ref{sec:factscore}), and describe how we created OpenFActScore from it (Section \ref{sec:openfactscore}). In Section \ref{sec:evaluation}, we evaluate a range of open-source models on the original FActScore data.

\section{FActScore}
\label{sec:factscore}
\subsection{Basics} 
FActScore \cite{min2023-factscore} is a model-based factuality measure. The computation of FActScore is based on so-called \textit{atomic facts}. Based on previous work \cite{nenkova-passonneau-2004-evaluating,shapira-etal-2019-crowdsourcing,zhang-bansal-2021-finding,liu-etal-2023-revisiting}, atomic facts are defined as short statements containing a single piece of information. A single sentence may contain multiple atomic facts. Since atomic facts are smaller than sentences, this allows for a more fine-grained analysis, avoiding subjective partial support labels \cite{min2023-factscore}.

FActScore makes three basic assumptions:
\begin{itemize}
    \item Whether an atomic fact is supported by a knowledge source has to be undebatable;
    \item Every atomic fact has an equal weight of importance, following  guidelines from \citet{krishna-etal-2023-longeval};
    \item Pieces of information in the knowledge base do not conflict or overlap with each other;
\end{itemize}

Based on these assumptions, FActScore splits the measurement of factuality into two steps:

\begin{enumerate}
    \item Atomic Fact Generation (AFG): Extracting individual atomic facts from a model output.
    \item Atomic Fact Validation (AFV): Verify for each claim whether it is supported by a reference knowledge source.
\end{enumerate}

After obtaining the Atomic Facts and validating them against  a knowledge source, the FActScore for a model is calculated as the number of Atomic Facts supported by the knowledge source divided by the total number of generated Atomic Facts.\footnote{This means that FActScore is essentially a precision-based measure. Recall is not measured automatically.}

\begin{figure}
    \centering
    \includegraphics[width=\linewidth]{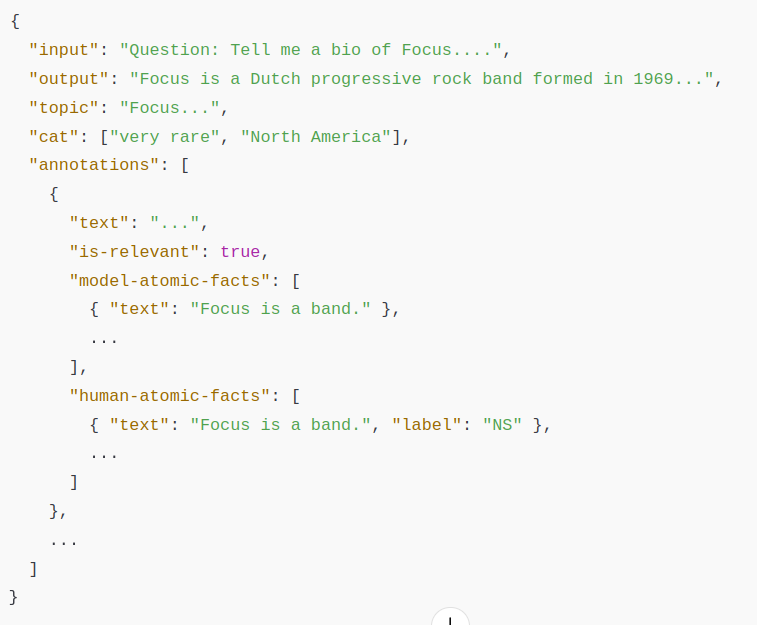}
    \caption{Sample from human annotated data. }
    \label{fig:fs_sample}
\end{figure}

\subsection{Human-Annotated Data} 
In the original paper, biography writing is used as an evaluation task fulfilling all the above requirements, as biographies are non-subjective and contain non-vague information. Similarly, it is argued that Wikipedia provides sufficient coverage of topics to also satisfy the prerequisites to be used as a knowledge base. The task is simply to write a biography about a given entity.

\citet{min2023-factscore} selected 183 Wikipedia entities to be human annotated for evaluation. During the annotation, human annotators mirrored the FActScore setup manually, first breaking down sentences into Atomic Facts, and second, verifying if they are supported by a knowledge source (here Wikipedia). In order to speed up the process, the authors produced biographies of three models to be evaluated and broke down the biographies into atomic facts using InstructGPT. Human annotators then just manually corrected and complemented the atomic facts, subsequently verifying their validity. This results in three sets of data, each one based on one of the models that were evaluated by the authors, namely InstructGPT \cite{ouyang2022training}, chatGPT\footnote{\url{https://chatgpt.com/}} and PerplexityAI\footnote{\url{https://www.perplexity.ai/}}.

Figure \ref{fig:fs_sample} shows an annotated sample.
The fields for \texttt{"input"} correspond to how the language model $LLM_{subj}$ was prompted, \texttt{"output"} to its corresponding response, and \texttt{"topic"} is what Wikipedia entity the model was prompted on.
The \texttt{annotation} field contains a list of annotation dicts, where \texttt{"text"} correspond to the sentence from which the Atomic Facts are being derived from.
If human annotators assign \textbf{False} to \texttt{is-relevant} the sentence is thus skipped without a validation step.
If the fact is relevant, they verify it using the English Wikipedia and label it as either \texttt{Supported} or \texttt{Not-supported}.

\subsection{Model}
The authors propose to fully automate this process using an \textit{estimator} model. The estimator first uses InstructGPT to decompose sentences into
Atomic Facts. Atomic facts are then validated by a different model, refereed to as $LM_{eval}$, which assigns \textbf{Supported} or \textbf{Not-Supported} labels depending on the text retrieved from a knowledge base.
When validating Atomic Facts, authors show the best validating strategy to be prompting $LM_{eval}$ by appending documents retrieved from a knowledge base to a simple prompt of "<atomic-fact>. True or False?". 
This approach is paired with a non-parametric approach, where individual tokens are systematically masked and a non-parametric masked language model \cite{Min_Shi_Lewis_Chen_Yih_Hajishirzi_Zettlemoyer_2023} is used to estimate likelihood.
The scores are then averaged across all tokens, and a final prediction is made by applying a predefined threshold.
The authors suggest an ensemble of both retrieval with prompting and the non-parametric prediction, where \texttt{Supported} \textbf{(S)} is only assigned in case both agree, otherwise they are assigned \texttt{Not-Supported} \textbf{(NS)}.


\subsection{Implementation}

The original FActScore implementation provides two configurations for estimating FActScore. In both, AFG is conducted using InstructGPT. Two different settings are then employed for AFV: One setting uses ChatGPT for validation, while the other employs a combination of a LLaMA model fine-tuned on Super-NaturalInstructions \cite{touvron2023llama, wang-etal-2022-super} and a non-parametric model to assess the validity of the Atomic Facts.

Technically, model generations are first broken down into individual sentences through simple parsing and text processing. Next, a few-shot prompting setup is constructed,  using a BM25 retriever to fetch semantically related samples of Atomic Facts that are combined into a prompt. This prompt is then used to query a model that performs Atomic Fact Generation (AFG), producing up to 128 new tokens. 
For each resulting Atomic Fact, relevant passages are retrieved from a Wikipedia dump by querying a Generalizable T5-based Retriever (GTR, \citet{ni-etal-2022-large}) with the combination of the original topic and the Atomic Fact.
The top five most relevant passages are prepended to a prompt for the validation language model ($LM_{eval}$), followed by the Atomic Fact to be validated, and the question “True or False?”, as described above. The model's output is then parsed for the first occurrence of either \texttt{"True"} or \texttt{"False"}. Finally, the number of AFs labeled as "True" is counted and FActScore is calculated as the proportion of supported Atomic Facts over the total number generated.

\section{OpenFActScore}
\label{sec:openfactscore}

\begin{table*}[t]
    \centering
    \begin{tabular}{ll|ccc|l}
        \toprule
        &&\multicolumn{3}{c|}{\textbf{original} $\mathbf{LLM_{subj}}$}\\
        && \textit{InstructGPT} & \textit{ChatGPT} & \textit{PPLAI} & \textit{Average} \\
        \midrule
        \multirow{4}{*}{\rotatebox[origin=c]{90}{\textbf{Evaluator}}}
        &\textit{Llama3.1} & 0.682 & \textbf{0.888} & \textbf{0.867} & 0.813 \\
        &\textit{Gemma}    & \textbf{0.878} & 0.870 & 0.852 & \textbf{0.867} \\
        &\textit{Qwen}     & 0.617 & 0.623 & 0.617 & 0.619 \\
        &\textit{Olmo}     & 0.873 & 0.878 & 0.840 & 0.864 \\
        \bottomrule
    \end{tabular}
    \caption{Average BERTScore F1 between model-generated atomic facts with 4 evaluator models and human-corrected atomic facts based on 3 ${LLM_{subj}}$ models. the maximum per column is printed in \textbf{bold}.}
    \label{table:bertscore}
\end{table*}

\subsection{Overview}
The original FActScore pipeline has limitations that hinder its broader adoption.
To reduce reliance on expensive proprietary models like InstructGPT and ChatGPT, the original implementatoin already used an alternative estimator, based on a LLaMA model fine-tuned on Super-NaturalInstructions \cite{touvron2023llama, wang-etal-2022-super} for Atomic Fact Validation. However, Atomic Fact Generation in their setup remains dependent on InstructGPT. Moreover, the original LLaMA weights are no longer easily accessible, having been superseded by newer versions. To address these limitations, we extend FActScore to support any model available on Hugging Face, making factuality evaluation more accessible and cost-effective. The result is OpenFActScore, a fully open and flexible alternative to the original framework. In this section, we discuss the main changes we made to FActScore to create our OpenFActScore.

Our implementation is based on the original FActScore codebase found at \url{https://github.com/shmsw25/FActScore}.
Text processing is implemented in multiple steps to ensure proper formatting for model input and to obtain Atomic Facts.
There also is an abstention detection module, which parses the model output to remove any output not directly relevant to the task. Model loading and prompting is done via classes developed by the authors.

Our implementation focuses on changes in the Atomic Fact Generation (AFG) and Atomic Fact Validation (AFV) steps.
The main change is the refactoring of one of the oringial model classes to be able to load models from HuggingFace, allowing support for chat templates and consequently, system prompts.

\subsection{HFModel Class}

Based on the code and architecture provided by \cite{min2023-factscore} we define a python class \texttt{HFModel}, capable of loading and prompting models from Hugging Face.
By leveraging the classes \texttt{AutoModelForCausalLM} and \texttt{AutoTokenizer} from the Transformers python library \cite{DBLP:journals/corr/abs-1910-03771} we are able to load the models on HuggingFace.
We implement flags for task selection, i.e. AFG or AFV, and  flags for selecting different validation strategies (retrieval, non-parametric, logits).

As we are loading models from HF we adapt the class to handle chat templates in conformity with best practices for prompting. This also allows us to generalize by using the templates provided by model owners.

\subsection{Chat Template and System Prompts}

Since OpenAI models and the original Llama do not use chat templates or system prompts we decided that implmenting them would help guide model production, improving final results on each task, AFG, and AFV.

The Atomic Fact Generation system prompt is constructed by listing simple directions and providing a \texttt{<demo>}, obtained from a list of human generated Atomic Facts provided by the authors.
The selected demo is the most relevant one based on the results of BM25 ranking between \texttt{<sentence>} and the available demos.
This is built the same way the original, but passed as System prompt, while originally it was passed as user input.

The Atomic Fact Verification prompt is simpler, as it provides only general instructions to guide model generation and suppress verbose answers.
The prompts can be seen in appendix \ref{sec:appendix}.

\section{Evaluation}
\label{sec:evaluation}



In this section, we evaluate the AFG and AFV modules based on diverse open-source models against the human-annotated gold standard.

We perform experiments with Gemma \cite{team2025gemma}, Qwen \cite{DBLP:journals/corr/abs-1910-03771}, and Llama 3.1-Instruct \cite{grattafiori2024llama3herdmodels}, which are currently popular open-usage models, and Olmo\cite{olmo20252olmo2furious}, which is a fully open source model.
In the following sections we describe the evaluation procedures in detail.

\subsection{Atomic Fact Generation}

\begin{table*}[t]
    \centering
    \begin{tabular}{ll|rr|rr|rr|c}
        \toprule
        &&\multicolumn{6}{c|}{\textbf{original} $\mathbf{LLM_{subj}}$}&\\
        &&\multicolumn{2}{c}{\textit{InstructGPT}} 
        &\multicolumn{2}{c}{\textit{ChatGPT}} 
        &\multicolumn{2}{c|}{\textit{PPLAI}} 
        &\multicolumn{1}{c}{\textit{Cumulative}} \\
        && ER & FS & ER & FS & ER & FS & \multicolumn{1}{c}{\textit{ER}} \\
        \midrule
        \multirow{5}{*}{\rotatebox[origin=c]{90}{\textbf{Evaluator}}}
        &\textit{Human}     &     0  & 42.5 &   0    & 58.3 &    0   & {71.5} &       \\
        &\textit{Llama3.1}  & -2.8  & 45.3 &  1.6  & 56.7 &  9.0  & 62.5 & 13.4  \\
        &\textit{Gemma}     & -6.3  & {48.8} & -1.7 & 60.0 &  4.2  & 67.3 & 12.2  \\
        &\textit{Qwen}      & -0.3  & 42.8 &  8.6  & 49.7 & 18.9  & 52.6 & 27.8  \\
        &\textit{Olmo}      & \textbf{-24.6} & 67.1 & \textbf{-20.1} & {78.4} & \textbf{-10.9} & 82.4 & \textbf{55.6} \\
        \bottomrule
    \end{tabular}
    \caption{Error Rate (ER) and FactScore (FS) using different models as Evaluator for atomic fact validation, evaluated against human references from three $\mathbf{LLM_{subj}}$ models. \textbf{Bold} indicates the ER with the highest magnitude per column.}
    \label{tab:ER_afv}
\end{table*}

To evaluate AFG, authors relied on human annotators' best judgment to write Atomic Facts on their own, or to rewrite the Atomic Facts generated by InstructGPT based on the output of three models, InstructGPT, chatGPT and PerplexityAI.

Since we do not have access to human annotators, we use the BERTScore-F1 \cite{bert-score} measure to estimate how semantically related the generated Atomic Facts are to the gold standard for each of the three models.
We compare the machine generated Atomic Facts to every human annotated Atomic Fact, for each sentence, and select the best match as final score. We then take the average of the best matches as measure of semantic overlap.

Table \ref{table:bertscore} presents the average BERTScore F1 between model-generated Atomic Facts and human-corrected Atomic Facts across four evaluators (Llama3.1 , Gemma, Qwen and Olmo). The human annotated Atomic Facts were obtained from productions from the $LLM_{subj}$ models InstructGPT, chatGPT and PerplexityAI, as mentioned above.

Gemma and Olmo show consistently strong semantic alignment with human AFs, across productions from all models, achieving average BERTScores of 0.867 and 0.864 respectively. This suggests both models are proficient at generating factually aligned content.

Llama3.1 performs well on ChatGPT and PPLAI evaluations (0.888 and 0.867), but its lower overlap with InstructGPT (0.682) worsens the average performance to 0.813. 

Qwen, on the other hand, underperforms across all evaluators, particularly with InstructGPT and PPLAI (0.617 for both), yielding the lowest overall score (0.619). On a closer look, we noticed that Qwen is apparently trained to "think", and to explain its own thinking process, which leads to increased verbosity and misalignment between human and model annotations.

Overall, the results suggests that Gemma and Olmo produce the most human-aligned Atomic Facts, while Llama3.1 is more evaluator-sensitive. Qwen would benefit from a better suited prompting strategy, which we leave for future work.

\subsection{Atomic Fact Validation}

To evaluate the atomic fact validation capabilities of diverse models, we use the AFV annotations provided by \cite{min2023-factscore} to evaluate models on their ability to validate Atomic Facts against the retrieved reference documents.

We report Error Rate in Table \ref{tab:ER_afv}, defined as the difference between human annotated FActScore and machine estimated FActScore, as a metric to indicate the differense to a human-estimated FActScore. Since each set of annotations has its intricacies, depending on the ${LLM_{subj}}$ used, we also report Cumulative ER, i.e. the summation of error rates, to give an intuition which model performs best overall on AFV. 
While we provide a validation method by comparing the logits, all of our experiments models use output parsing as a mechanism for detecting validation response as \texttt{True} or \texttt{False}.

The \textit{human} row provides reference FactScores (FS) for each subject. Negative ER values indicate that a model acting as Evaluator undershoots FactScore, while a positive value indicates overestimation.

Olmo shows the largest deviations from human judgments, with notably high ERs for InstructGPT (+24.6), ChatGPT (+20.1), and a negative ER for PPLAI (-10.9), indicating inconsistent alignment and highest cumulative ER (55.6).
Despite performing well during AFG (as indicated by a high BERTScore F1 in Table \ref{table:bertscore}), this suggests Olmo’s factual verification may not reflect human judgments reliably.

Qwen performs inconsistently, particularly with ChatGPT and PPLAI (ERs of +8.6 and +18.9), leading to a cumulative ER of 27.8, confirming its previously observed struggles.

Llama3.1 and Gemma both display relatively low cumulative ERs (13.4 and 12.2), suggesting better alignment with human evaluations. Llama3.1 slightly overestimates FactScore under PPLAI, while Gemma performs more consistently across evaluators.

In summary, despite Olmo’s high generation quality, it results in large deviations when used as an evaluator. Conversely, Gemma and Llama3.1 emerge as more reliable evaluators, offering closer approximations to human judgments in Atomic Fact Validation.

Based on the results presented on Table \ref{table:bertscore} and Table \ref{tab:ER_afv}, using Gemma as $LM_{eval}$ for Atomic Fact Validation and Olmo for fact generation seems to be the most reliable option. Although Olmo presented lower performance as compared to Gemma, this difference is negligible, 0.35\%. Since Olmo is completely open source, providing sources for data used for training, architecture and training recipes, we propose to use Olmo for Atomic Fact Generation.

\subsection{Overall Evaluation}
\begin{table}[h]
    \centering
    \begin{tabular}{l|r|r|r}
        \toprule
        \textbf{Model} & \textbf{FS (A)} & \textbf{FS (B)} & \textbf{FS (OFS)} \\
        \midrule
        GPT-4                & \textbf{73.1} & 59.9 & \textbf{50.08} \\
        ChatGPT              & 71.6 & \textbf{60.4} & 46.52 \\
        Alpaca 65B           & 55.6 & 46.3 & 37.14 \\
        InstructGPT          & 52.8 & 41.7 & 35.89 \\
        Alpaca 13B           & 47.7 & 40.3 & 30.03 \\
        Vicuna 7B            & 38.9 & 36.9 & 29.05 \\
        MPT Chat 7B          & 30.1 & 27.9 & 20.65 \\
        Pythia 12B     & 25.1 & 20.8 & 16.23 \\
        Dolly 12B            & 21.7 & 17.1 & 13.45 \\
        StableLM 7B    & 17.3 & 16.3 &  9.20 \\
        \bottomrule
    \end{tabular}
    \caption{Comparison of factual performance across models and for different settings. \textbf{FS (A)} and \textbf{FS (B)} are FActScores provided by \citet{min2023-factscore}, using InstructGPT+chatGPT (A) and InstructGPT+InstLlama+npm (B). \textbf{FS (OFS)} is FActScore as computed by a combination of Olmo for AFG and Gemma for AFV in OpenFActScore. Maximum values per column are printed in \textbf{Bold}.}
    \label{tab:factscore_comparison}
\end{table}
To evaluate our proposed setup we obtained FActScore for 10 Large Language models prompted for the Biography Writing task. These model productions are available in the original FActScore repository for download.
In Table \ref{tab:factscore_comparison}, we show the FActScore obtained from OpenFactScore (column \textbf{FS (OFS)}) along with the two settings (A) and (B) proposed by \citet{min2023-factscore}.

Even though the OpenFActScore scores are lower overall, the ranking of the models stays the same.
The Pearson correlation between FActScore and OpenFActScore in both setting (A) and (B) is over 0.99, showing that open models follow very similar performance patterns to closed models.

\section{Related Work on Factuality Evaluation}

Hallucinations and factuality remain open challenges in research \cite{george2023factoredverificationdetectingreducing, alkaissi2023artificial, zhang2025llmhallucinationspracticalcode}.
Methods for evaluating factuality vary widely, ranging from question-answering approaches \cite{durmus2020feqa}, to faithfulness classification datasets \cite{huang2020knowledge}, natural language inference (NLI) based evaluations \cite{falke-etal-2019-ranking}, and more recently popular, model-based methods \cite{filippova2020controlled, min2023-factscore}.

A key challenge in factuality evaluation lies in defining what constitutes a hallucination.
Common taxonomies distinguish between \textbf{intrinsic} hallucinations, which contradict the prompt, and \textbf{extrinsic} hallucinations, which introduce information not grounded in the prompt but potentially factual \cite{Ji_2023}.
Interestingly, extrinsic hallucinations may still be considered useful if they provide factual information \cite{cao-etal-2022-hallucinated}.


FActScore can be seen as a model-based evaluation method designed to detect intrinsic hallucinations. Althought a hallucination might not directly contradict the context, generations are evaluated against a knowledge source, which should be considered as the documents which it need to abide by.

\section{Conclusion}

In this paper we presented OpenFActScore, an open-usage version of FActScore.
We tested four of the currently relevant open-usage models in two tasks, Atomic Fact Validation and Atomic Fact Generation.
Our experiments reported that Gemma \cite{team2025gemma} had the best performance for Atomic Fact Validation by presenting the lowest cumulative Error Rate, and Atomic fact Generation by providing the best BERT F1 Score among the evaluated models.
We, however, chose \citet{olmo20252olmo2furious} as an Atomic Fact Generation model for being on par with Gemma (0.3\% difference), and also being fully open, providing not only weights and architechture but its training data.
We evaluated productions from 11 different language models and compared our FActScore with the original scores, and report a 0.99 Pearson correlation between the two settings provided in the repository of the original authors.





\bibliography{acl_latex}

\appendix

\section{System Prompts} \label{sec:appendix}

\subsection{Atomic Fact Generation}

\begin{lstlisting}
SYSTEM_PROMPT = You are an annotator that breaks down sentences into independent facts, short statements that each contain one piece of information contained in the given sentence. In the next paragraphs you have examples of sentences broken down in atomic facts. You have to complete the example given by the user. Do not add new entities, do not deviate from the subject of the sentence given by the user, do not hallucinate, do not repeat facts in the system prompt. List the sentences using -
                
Please break down the following sentence into independent facts: <demo>
\end{lstlisting}

\begin{lstlisting}
USER_INPUT = Please break down the following sentence into independent facts: <sentence>
\end{lstlisting}

\subsection{Atomic Fact Validation}

Next we provide the prompt for Atomic Fact Validation.

\begin{lstlisting}
SYSTEM_PROMPT = "You are an annotator that verifies the factuality of a sentence according to a given source text. You answer only True or False and provide no further explanations."

USER_PROMPT = "Input: "Answer the question about <entity> based on the given context.
              [Title: <document_title>]
              [Text: <document_content>]
              True or False?
              Answer:
"
\end{lstlisting}

\section{Demo}

A simple demonstration of our system can be found in: \url{https://youtu.be/zXdk3b4alrE}

\end{document}